\documentclass[sigconf, 10pt]{acmart}
\usepackage{graphicx}
\usepackage{subfig}
\usepackage{caption}
\usepackage{multirow}
\usepackage{multicol}
\usepackage{array}
\usepackage{booktabs}
\usepackage{stfloats}
\usepackage{footmisc}
\usepackage{float}
\usepackage{enumitem}
\usepackage{multirow}

\AtBeginDocument{%
  \providecommand\BibTeX{{%
    \normalfont B\kern-0.5em{\scshape i\kern-0.25em b}\kern-0.8em\TeX}}}

\setcopyright{acmcopyright}
\copyrightyear{2018}
\acmYear{2018}
\acmDOI{XXXXXXX.XXXXXXX}

\acmConference[Conference acronym 'XX]{Make sure to enter the correct
  conference title from your rights confirmation emai}{June 03--05,
  2018}{Woodstock, NY}
\acmPrice{15.00}
\acmISBN{978-1-4503-XXXX-X/18/06}

\author{Dhruv Patel, Ankita Kumari Jain, Haikoo Khandor, Xhitij Choudhary, Nipun Batra}
\email{{patel.dhruv, ankitajains, haikoo.ashok, xhitij.cm, nipun.batra}@iitgn.ac.in}
\affiliation{%
  \institution{Indian Institute of Technology, Gandhinagar}
  \city{Gandhinagar}
  \country{India}
}

\providetoggle{hide}
\settoggle{hide}{false}

\begin{document}

\title{Benchmarking Active Learning for NILM}

\begin{abstract}
Non-intrusive load monitoring (NILM) focuses on disaggregating total household power consumption into appliance-specific usage. Many advanced NILM methods are based on neural networks that typically require substantial amounts of labeled appliance data, which can be challenging and costly to collect in real-world settings. We hypothesize that appliance data from all households does not uniformly contribute to NILM model improvements. Thus, we propose an active learning approach to selectively install appliance monitors in a limited number of houses. This work is the first to benchmark the use of active learning for strategically selecting appliance-level data to optimize NILM performance. We first develop uncertainty-aware neural networks for NILM and then install sensors in homes where disaggregation uncertainty is highest. Benchmarking our method on the publicly available Pecan Street Dataport dataset, we demonstrate that our approach significantly outperforms a standard random baseline and achieves performance comparable to models trained on the entire dataset. Using this approach, we achieve comparable NILM accuracy with approximately ($\sim30\%$) of the data, and for a fixed number of sensors, we observe up to a 2x reduction in disaggregation errors compared to random sampling.

\end{abstract}




\maketitle

\section{Introduction}
Non-Intrusive Load Monitoring \cite{hart1992nonintrusive}, or (NILM), disaggregates the total power consumption into appliance-level usage. Prior studies \cite{lund2015review} suggest that knowing appliance-wise power consumption can help users reduce their energy consumption by $15\%$. 

Since George Hart's pioneering work on NILM \cite{hart1992nonintrusive}, several algorithms have been proposed, including but not limited to time series models such as additive factorial hidden Markov models \cite{kolter2012approximate}, discriminative sparse coding \cite{kolter2010energy}, graph signal
processing \cite{he2016non}. The first neural network based approach was proposed by Kelly et al. in 2015 \cite{kelly2015neural}. Many neural network based techniques have been proposed since then \cite{zhang2018sequence,10.1145/3486611.3486652,huber2021review, bansal2022not}.

Conventional neural network (NN) models generally work well when trained on a large labeled dataset. However, collecting labeled data via individual appliance power meters in every household is prohibitively expensive. But, in general, mains energy data is easier to acquire than appliance-level data. As a result, we have access to a huge number of houses that have smart meter data but no appliance-level instrumentation. For the rest of the paper, we refer to such houses as unlabeled houses. In this paper, we investigate the possibility of \emph{achieving accurate disaggregation performance by leveraging training data (appliance-level data) from a limited selection of strategically chosen houses drawn from a vast pool of unlabeled houses.}





This paper introduces, for the first time, an active learning framework for strategically selecting appliance-level data to enhance NILM performance, and we evaluate this approach in a comprehensive benchmarking setting. Active Learning \cite{settles2009active} is a subfield of machine learning that employs a selective approach to identify and label data that maximizes uncertainty reduction from a large pool of unlabeled examples. This methodology aims to minimize the cost associated with labeling or annotating data. An acquisition function determines the utility of a data example for improving the model performance.

In this paper, we use the state-of-the-art neural networks for Non-Intrusive Load Monitoring (NILM) models. We leverage the recent techniques for quantifying the uncertainty in model predictions \cite{van2020uncertainty, bansal2022not}. We specifically use Monte Carlo Dropout~\cite{gal2016dropout} to estimate uncertainty and two acquisition functions: entropy and mutual information~\cite{shannon1948mathematical}. We also discuss mechanisms to add the temporal context in our decision making.  

We benchmark our models on the publicly available Pecan Street Dataport dataset~\cite{7418187}. We  benchmark active learning in NILM, addressing the practical challenges of appliance data collection. We first evaluated the ideal setting: where we can add a sensor (appliance-level meter) to maximize the predictive performance of a single appliance, without considering other appliances. We found our approaches to work significantly better than the random baseline. However, we also found this ideal setting to be impractical, as the set of houses chosen for different appliances had low intersection. 

We further evaluated our approach on the realistic setting, where we jointly optimize for multiple appliances by picking houses to install sensors where the predictive performance of all appliances is likely to improve most. We found our approach to be better than the random baseline. For appliances like air conditioner and furnace, we are almost $2$x better than the random baseline in the initial iterations of active learning. We require a significantly lesser number of sensors, such as $50\%$ for air conditioner, $40\%$ for furnace and dishwasher as compared to using all the sensors in the random baseline to get a similar level of performance.

\section{Background and Related Work}


We now discuss the relevant background and related work.

\subsection{Neural Energy disaggregation}

As mentioned in the introduction, many neural network approach for energy disaggregation have been used and Seq2Point \cite{zhang2018sequence} model which uses $1$d CNN approach still continues to be one of the most popular approach. A sequence length is fed as an input and a point prediction of the appliance power series is taken as the output. 


\subsection{NNs and Uncertainty}


In the context of machine learning, we generally consider the following types of uncertainty:\\
\noindent \textbf{1) Aleatoric uncertainty}: which is the inherent uncertainty due to variability in the data generation process, noise in the data or measurement errors. Even with an infinite amount of data, aleatoric uncertainty cannot be reduced. \\
\noindent \textbf{2) Epistemic uncertainty}: which is the uncertainty arising from an incomplete understanding of the model's underlying parameters or structure. With more data or by improving the model architecture, we can minimise epistemic uncertainty.\\

\noindent We now briefly discuss different techniques for uncertainty quantification in neural nets. The recent surveys shows the in-depth discussions on these different methods \cite{gal2016dropout}\cite{van2020uncertainty}.

\subsubsection{Homoskedastic Regression}
Homoskedastic neural networks model the same variance across all inputs.
We assume our predictive distribution to be a Gaussian distribution with mean $\mu_x$ and a constant sigma ($\sigma$) for all $x$. Hence they do account for uncertainty but are not useful for active learning \cite{settles2009active}.

\subsubsection{Heteroskedastic Regression}
Heteroskedastic models learn different variance for different data points. Thus, heteroskedastic neural networks estimate both $\mu(x)$ and $\sigma(x)$ as a function of the input $x$ (via two output nodes in the final layer). These models only consider the data (aleatoric) uncertainty but do not account for the uncertainty in estimating the parameters (epistemic). 

\subsubsection{Bayesian Neural Networks}
\label{MC_dropout}
A Bayesian neural network puts a prior $p(\boldsymbol{\theta})$ on the model weights ($\boldsymbol{\theta}$), assumes a homoskedastic or heteroskedastic likelihood, and computes the posterior ($p(\boldsymbol{\theta}|\mathcal{D}_{\text{train}})$) i.e the distribution of model parameters having observed the data.

Bayesian Neural Networks (BNNs) accounts for both aleatoric and epistemic uncertainty, however, such an approach would be computationally intractable \cite{blundell2015weight} and hence we perform certain approximations to quantify the uncertainties.

\begin{itemize}
    \item Sampling based methods: Sample from probability distributions to approximate the intractable integrals. Markov chain monte carlo (MCMC) \cite{robert1999monte} involve constructing a Markov chain whose final posterior distribution serves as the chain's equilibrium distribution. Several MCMC based methods like NUTS \cite{No-U}, Metropolis Hastings \cite{hastings1970monte}, Hamiltonian Monte Carlo \cite{neal2011mcmc}, etc. have been proposed.
    
    \item Variational approximation: Variational inference \cite{haussmann2020sampling} approximates the true posterior $p(\boldsymbol{\theta}|\mathcal{D}_{\text{train}})$ by a proxy distribution $q_\phi(\boldsymbol{\theta})$ with a known functional form parameterized by $\phi$ and minimizes the Kullback-Leibler divergence (KL divergence) \cite{kullback1951information} between $q_\phi(\boldsymbol{\theta})$ and $p(\boldsymbol{\theta}|\mathcal{D}_{\text{train}})$, denoted as:
    \begin{equation*} \text{KL}\left(q_\phi(\boldsymbol{\theta})\|\|p(\boldsymbol{\theta}|\mathcal{D}_{\text{train}}\right).
    \end{equation*}
    After a few algebraic manipulations, minimizing this KL divergence and maximizing the functional below turn out to be equivalent:
    \begin{equation*}
        \begin{aligned}
            \mathcal{L}_{\text{ELBO}} &= \mathbb{E}_{q_\phi(\boldsymbol{\theta})}\left[\log p(y|\boldsymbol{\theta},\mathcal{D}_{\text{train}} )\right] \quad \text{(Data Term)} \\
            &\quad - \text{KL}\left[q_\phi(\boldsymbol{\theta})\|\|p(\theta)\right] \quad \text{(Regularization Term)}.
        \end{aligned}
    \end{equation*}

This functional is often referred to as the Evidence Lower Bound (ELBO). 

    \item Laplace approximation \cite{tierney1986accurate}: Uses a normal approximation centered at the maximum of the posterior distribution, or maximum a posteriori (MAP).
    
    \item Dropout as Bayesian approximation: A much more computationally efficient approach, MC Dropout \cite{gal2016dropout} has been introduced.  It is one of the popular techniques for efficiently estimating uncertainty in neural networks. In MC dropout, random dropout of nodes occurs during prediction. This dropout behavior is governed by a Bernoulli distribution, determining the probability of a node being either retained or dropped. It is worth noting that MC Dropout bears similarity to regular dropout, a method utilized to mitigate model overfitting \cite{srivastava2014dropout}. 
    Post training, we run $F$ forward passes of our trained model with nodes dropped as per Bernoulli distribution. For each of the $F$ passes, we predict the mean ($\mu_{x}$) and sigma ($\sigma_{x})$ for an input $x$. We weigh each Gaussian obtained from the forward pass in the mixture of Gaussians uniformly \cite{lakshminarayanan2017simple}. We can thus write the predictive distribution as:
    \begin{equation}\label{mc1}
        p(y|x, \mathcal{D}_{\text{train}}) =\frac{\sum_{i=1}^F \mathcal{N}\left(x \mid \mu_i(x), \sigma_i(x)\right)}{F}
    \end{equation}
    Or, we can write that the predictive distribution as a Gaussian mixture whose mean and standard deviation is given by:
    \begin{equation} \label{mc2}
    \mu_\textrm{ensemble}(x)=\frac{\sum_{i=1}^{F}\mu_i(x)}{F} 
    \end{equation}
    \begin{equation}\label{mc_sigma_eqn}
        \sigma_\textrm{ensemble}(x)= \sqrt{\frac{\sum_{i=1}^{F}(\sigma_i(x)^2+\mu_i(x)^2)}{F}-\mu_\textrm{ensemble}(x)^2}
    \end{equation}
\end{itemize}

\subsection{Active Learning}
\begin{figure}[htp!]
\includegraphics[scale=0.1]{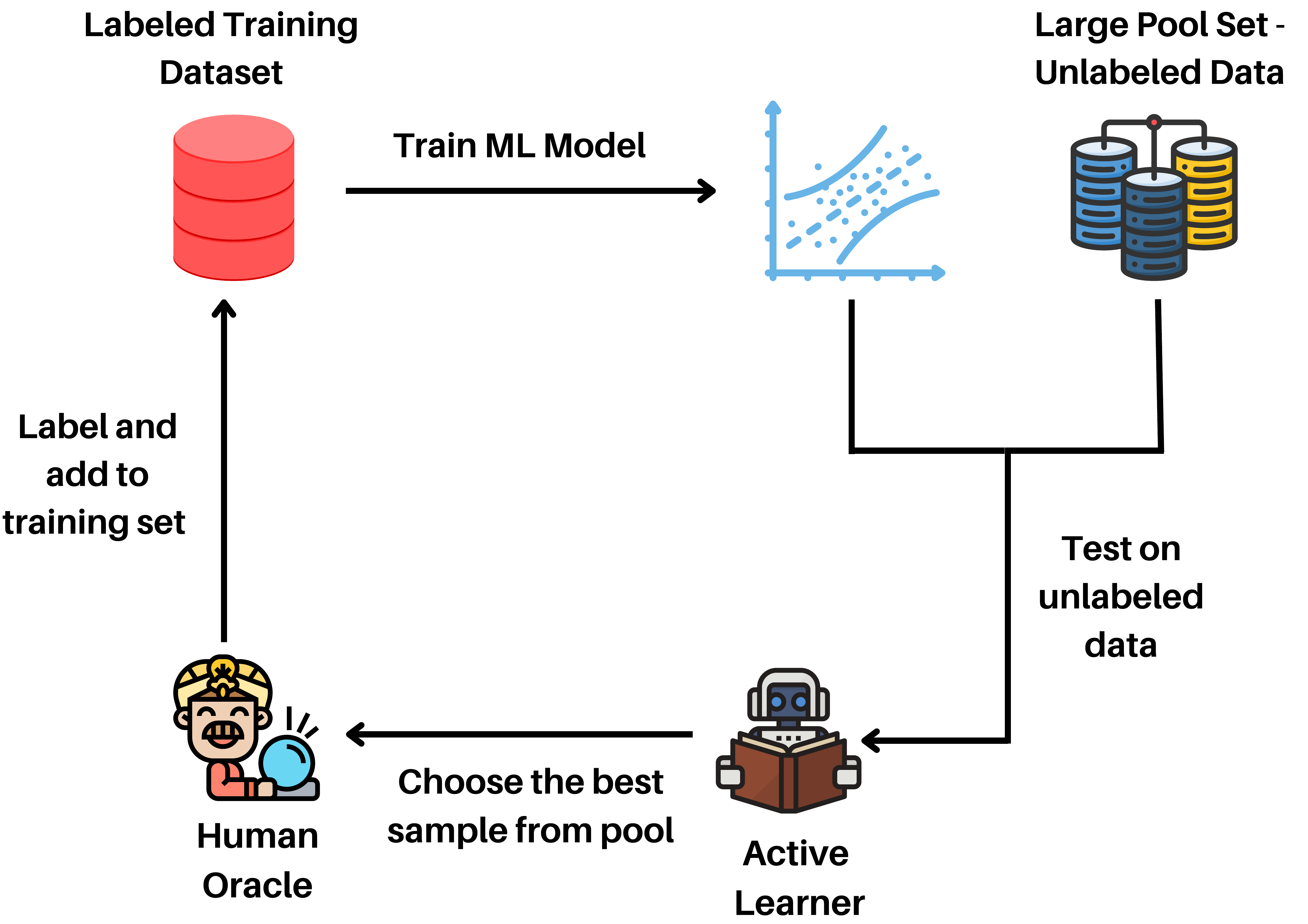}
    \caption{Active learning (AL) loop showing the flow of data across the whole AL pipeline}
\label{fig:AL-loop}
\end{figure}
\noindent Our approach uses active learning paradigms to overcome bottlenecks while achieving the best performance with fewer labeled samples. A quintessential active learning framework consists of four major parts:
\begin{enumerate}
    \item a small labeled dataset
    \item an active learner
    \item an expert Oracle
    \item unlabeled data pool
\end{enumerate}


In the active loop, we first train our model on a small labeled training dataset, then we actively query the data points from a large unlabelled pool dataset, label them and keep expanding our training dataset with these queried data points. Figure \ref{fig:AL-loop} illustrates the AL loop. 

The active learner proactively queries data it wants to learn from instead of being passively trained on all the available data. It leverages uncertainty in predictions to pick out the best points from the unlabeled data that significantly increase model performance. 

In traditional ML models, interactive querying of pool points is done via various uncertainty acquisition functions. For example: Query by Committee , Entropy Sampling, and Maximum Standard-Deviation based sampling \cite{settles2009active}. The query equation is given by 
\begin{equation} \label{query}
\hat{\boldsymbol{x}}_*=\operatorname{argmax}_{x \in x_{\text {pool }}} A(\boldsymbol{x})
\end{equation}
where  $A\left(\boldsymbol{x}\right)$ is the acquisition function and $\hat{\boldsymbol{x}}$ is the queried point from the pool set. 


\subsection{Active Learning for NILM}
We propose using active learning \cite{settles2009active} for reduction in labeled data requirement for low frequency domain in NILM.



There is limited prior work on active learning for NILM. Previous literature like \cite{TODIC2023121078}, use Wave-Net \cite{jiang2021deep} and modified BatchBALD \cite{kirsch2019batchbald} for NILM task and \cite{9121244} use stream based uncertainty sampling where the notion of time series is not upheld. Both of the papers also attempt to solve a classification task. They adopt a user-based querying approach. Users are prompted to use an app or given the snapshot/time-series showing the aggregate power consumption time series, for example, $11:00-11:30$ am and then asked what appliance they were using at that time. \textbf{To the best of our knowledge we are the first to benchmark and apply active learning for NILM framework in a regression setting.} We propose an active learning approach based on installing appliance-level sensors. We query a house and add a sensor to the queried house. Moreover, we use pool-based uncertainty sampling for time series for a regression task. Our pool-based sampling approach makes our active learning settings more practical as compared to the stream-based samplings performed in prior literature \cite{9121244}. 


\section{Problem Statement and Notation} 
In this research paper, we focus on developing a practical and data-efficient Non-Intrusive Load Monitoring (NILM) approach that achieves performance comparable to conventional NILM methods at a significantly lower cost. We aim to accomplish this by leveraging a set of candidate/pool houses willing to have sensors installed, along with training houses. We now discuss the different kinds of houses in our problem setup. \\
\noindent \textbf{1. $\text{H}_{\text{train}}$} denotes the set of train houses, for which each house $h$ in $H_{\text{train}}$ is associated with an aggregate (\textbf{M}ains) power time series $P_{h}^M(t)$, where $t$ represents the time index. For  $H_{\text{train}}$ we always have access to both, the aggregate power time series $P_{h}^M(t)$ and the appliance power time series $P_{h}^a(t)$.

\noindent \textbf{2. $\text{H}_{\text{pool}}$} denotes the set of pool houses, which are the potential sources for querying appliance power data whose ground truth appliance data $P_{h}^a(t)$ becomes available only once the house is queried. Their aggregate power time series $P_{h}^M(t)$ is available for the entire duration.

\noindent \textbf{3. $\text{H}_{\text{test}}$} represents the set of test houses, which are unseen houses that serve as a validation or evaluation set. We do not have access to their aggregate power or appliance power time series during the active learning process. However during testing, we have access to the aggregate/mains power  reading of the test houses. 

\noindent \textbf{4. }The list of pool, train and test houses are separate and no house is shared between the three lists.\\


Our goal is to select a specific pool house $h_p$ and obtain its appliance power time series $\{P_{h_p}^a(t_1:t_2)\}$ from time $t_1$ till the end time of training, $t_2$. \\
$P$ - Power. \\
$h$ - particular house. \\
$a$ - particular appliance out of $a_1, a_2, \ldots, a_n$.\\
$M$ - Mains power reading. \\
$t_1:t_2$ - Time $t_1$ to time $t_2$. 

Since it is a time series problem setup, we need to devise a strategy to make effective use of the available temporal data and make inferences for future time steps. Assume the initial train and pool sets to be:

\noindent Train houses: $H_{\text{train}} = {h_1, h_2, h_3}$. \\
Pool houses: $H_{\text{pool}} = {h_5, h_6, h_7, h_9}$. 

\begin{figure*}
  \centering
      \includegraphics[scale = 0.72]{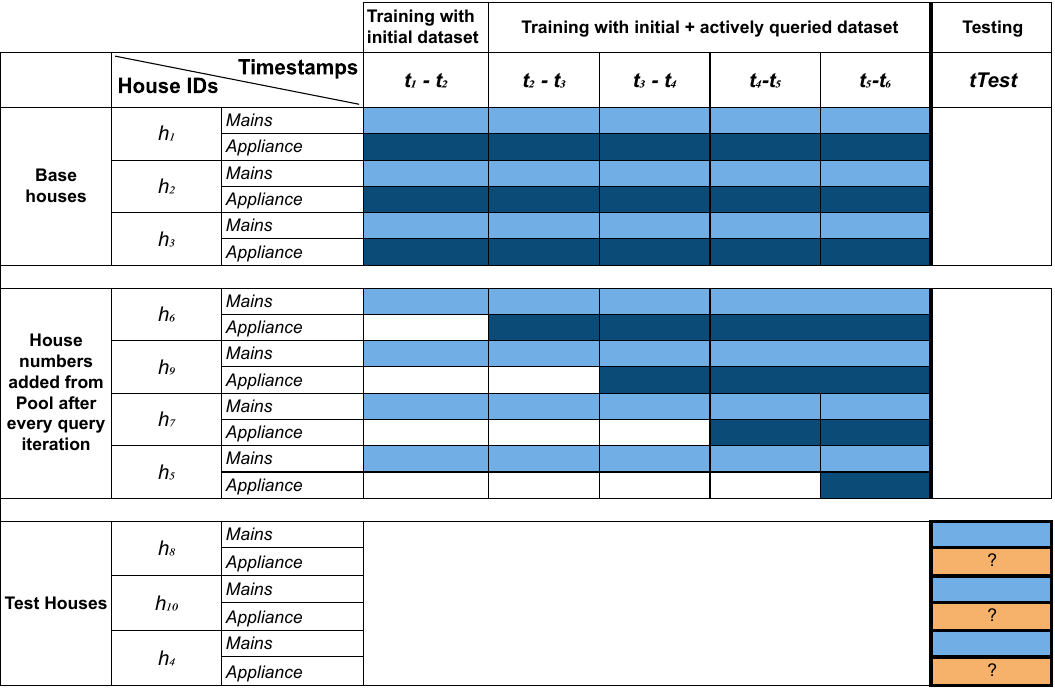}
  \caption{The mains power data is available for all the houses in the experiment (light blue color). The appliance power (dark blue color) used for training the model is available only when the house is added to the training set. For the test houses, we predict the appliance power denoted by a "?" mark and validate our performance.}
  \label{problem_statement}
\end{figure*}

\noindent As per figure \ref{problem_statement}, we have the mains power for train houses and pool houses from time $\textrm{t}= t_{1}$ to $\textrm{t} = t_{n} $. However the ground truth for the entire duration is available only for the train houses. 

\noindent At time $t = t_{2}$, we strategically select a pool home $h_6$, remove it from the pool set and add it to the train set. The updated sets would then be:

\noindent Train houses: $H_{\text{train}} = {h_1, h_2, h_3, h_6}$. \\
Pool houses: $H_{\text{pool}} = {h_5, h_7, h_9}$.\\
By selecting a new pool home at regular intervals, we can continuously expand the train set and incorporate new labeled data, allowing the model to learn from a larger and more diverse dataset. We then test our model's performance on the test houses ($H_{\text{test}}$).




\section{Approach}




Our objective is to quantify uncertainty and using less but informative data to train the model. We begin by addressing our practical setting, examining the pool dataset, and concluding by explaining the single-task and multi-task settings. 
\subsection{Disaggregation Method}

\subsubsection{Single Output}
In single output models, we train a separate model for each appliance and also set the hyperparameters specific to the appliance. We use current state-of-the-art sequence to point (S2P) \cite{zhang2018sequence} neural networks for NILM framework \cite{bansal2022not,batra2019demonstration} as the model. We use S2P with heteroskedastic likelihood. Our output is a Gaussian or Normal distribution. Thus, we predict $\mu(x)$ and $\sigma(x)$ per appliance for all input $x$. Figure \ref{fig:s2p_archs}a shows the model architecture, where the input is a sequence of aggregate power and output, $\mu(x)$ and $\sigma(x)$  is disaggregated power consumption of an appliance for the mid-point of the input time sequence.



\subsubsection{Multi Output}
\begin{figure}[h]
\centering
\includegraphics[width=0.9\columnwidth]{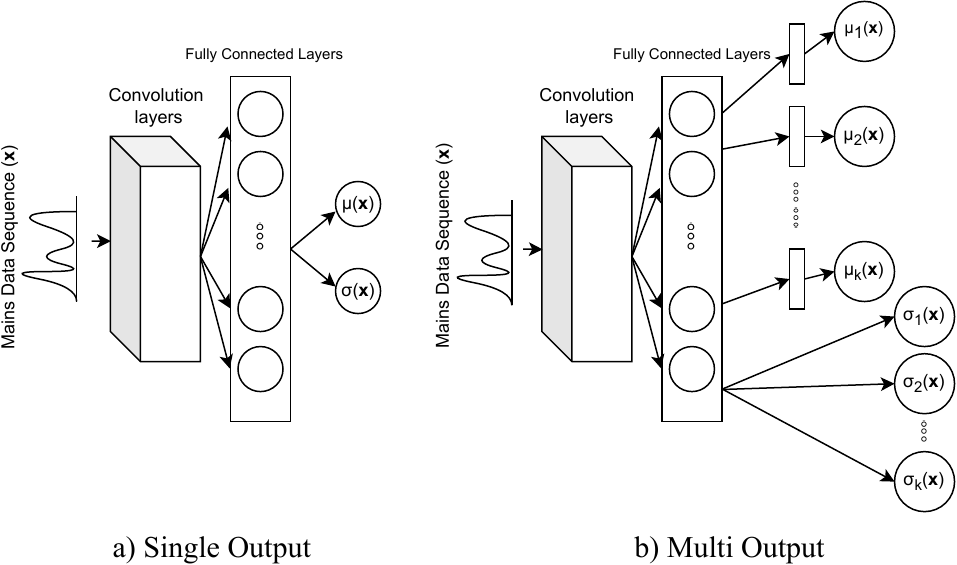}
\caption{Seq2Point architecture of a) Single Output and b) Multi Output}
\label{fig:s2p_archs}
\end{figure}

In multi output models, we train a single model that predicts the mean and sigma for all appliances. This accounts to a total output of $2k$ parameters where $k$ is the number of appliances. We find learning the sigmas first and then a separate layer for learning means to be better than learning all the $2k$ parameters in the final layer. Figure \ref{fig:s2p_archs}b shows the model architecture. 

\subsection{Acquisitions functions}
We query the pool houses based on two different acquisition strategies namely entropy and mutual information. We use MC-Dropout as the strategy to quantify the uncertainty by running $N$ stochastic forward passes (with dropout) of the trained network as discussed in details in Section~\ref{MC_dropout}.
\subsubsection{Entropy}
We use differential entropy as an estimate to quantify uncertainty in our model \cite{garbaczewski2006differential}. For a univariate Gaussian, the expression for entropy~\cite{shannon1948mathematical} for a general random variable ${X}$ is give as:
\begin{equation}
    H(X) = \log (\sqrt{2\pi e}\sigma)
\end{equation}
We observe that the entropy is directly proportional to the standard deviation. Hence, we can approximate the entropy of the predictive distribution (up to a constant) based on equations \ref{mc1},~\ref{mc2} and \ref{mc_sigma_eqn} as: 
\begin{equation}\label{entropy_eqn}
\mathrm{A_{Entropy}}(x) =  \mathbb{H}\left[y \mid \mathbf{x}, \mathcal{D}_{\text {train }}\right] \approx \sigma_\mathrm{{ensemble}}(\mathbf{x})
\end{equation}
\subsubsection{Mutual Information}\label{mutual information}

Mutual information quantifies the reduction in uncertainty of one variable ($X$) given information about another variable ($Y$). It is generally evaluated using $H(X;Y) = H(X) - H(X|Y)$
where $H(X)$ is the marginal entropy of $X$ and $H(X|Y)$ is the conditional entropy of $X$ given $Y$.

As per mutual information acquisition, we want to choose pool points that are expected to maximize the information gained about the model parameters, i.e. maximise the mutual information between predictions ($X$) and model posterior ($Y$)
\begin{align}\label{mi_eqn}
\mathrm{A_{MI}}(x) &= \mathbb{I}[y, \boldsymbol{\theta} \mid \mathbf{x}, \mathcal{D}_{\text{train}}] \nonumber \\
&= \mathbb{H}[y \mid \mathbf{x}, \mathcal{D}_{\text{train}}] - \mathbb{E}_{p(\boldsymbol{\theta} \mid \mathcal{D}_{\text{train}})}[\mathbb{H}[y \mid \mathbf{x}, \boldsymbol{\theta}]]
\end{align}

\noindent with $\boldsymbol{\theta}$ the model parameters (here $\mathbb{H}[y \mid \mathbf{x}, \boldsymbol{\theta}]$ is the entropy of $y$ given model weights $\boldsymbol{\theta}$ ).

While calculating the acquisition function for entropy ($\mathrm{A_{Entropy}}(x)$) as per equation \ref{entropy_eqn}, we could approximate it to a constant, but for calculating the acquisition function for mutual information ($\mathrm{A_{MI}}(x)$ ) as per equation \ref{mi_eqn}, we cannot make a similar approximation. The closed form solution of $\mathrm{A_{MI}}(x)$ is intractable \cite{kirsch2019batchbald} and hence we approximate it. 

To estimate the $\mathbb{H}[y \mid \mathbf{x}, \mathcal{D}_{\text{train}}] $, we use Monte Carlo sampling given by equation \ref{predictive_entropy}.
\begin{equation}\label{predictive_entropy}
 \mathbb{H}[y \mid \mathbf{x}, \mathcal{D}_{\text{train}}]  \approx -\frac{1}{S} \sum_{s=1}^{S} \log \left( \frac{1}{F} \sum_{i=1}^{F} p_i(\mathbf{x}_s) \right) 
\end{equation}
\noindent where,  \( S \) is the number of Monte Carlo samples drawn from the mixture of Gaussians (MOG), $F$ is the number of Gaussians in the MOG, \( \mathbf{x}_s \) represents the \( s \)-th sample drawn from the MOG and 
\( p_i(\mathbf{x}_s) \) is the probability of \( \mathbf{x}_s \) belonging to the \( i \)-th Gaussian component. To estimate \( p_i(\mathbf{x}_s) \), we sample \( \mathbf{x}_s \) from the \( i \)-th Gaussian component using its mean \( \mu_i \) and standard deviation \( \sigma_i \) as follows:
\begin{equation}
    \mathbf{x}_s \sim \mathcal{N}(\mu_i(x_s), \sigma_i(x_s)^2) 
\end{equation}

Then, \( p_i(\mathbf{x}_s) \) is calculated using the probability density function of the normal distribution:
\begin{equation}
 p_i(\mathbf{x}_s) = \frac{1}{\sqrt{2 \pi \sigma_i(x_s)^2}} \exp \left( -\frac{(\mathbf{x}_s - \mu_i(x_s))^2}{2 \sigma_i(x_s)^2} \right) 
\end{equation}

Now, to estimate the $\mathbb{E}_{p(\boldsymbol{\theta} \mid \mathcal{D}_{\text{train}})}[\mathbb{H}[y \mid \mathbf{x}, \boldsymbol{\theta}]]$, we use Equation \ref{expected_entropy}.  
\begin{equation}
\mathbb{E}_{p(\boldsymbol{\theta} \mid \mathcal{D}_{\text{train}})}[\mathbb{H}[y \mid \mathbf{x}, \boldsymbol{\theta}]] = \sum_{i=1}^{F}\frac{1}{2}\log_{e}(2\pi\sigma_{i}^2) \label{expected_entropy}\\ 
\end{equation}
Finally, we estimate, $\mathrm{A_{MI}}(x)$ as per equation \ref{mi_eqn} by taking the difference of quantities in Equation \ref{predictive_entropy} and \ref{expected_entropy}.

\textbf{It is to be noted that differential entropy takes into account both the aleatoric as well as epistemic uncertainty whereas mutual information takes into account only the epistemic uncertainty \cite{gal2017deep}.}
\subsection{Active Learning settings}






We employ an active learning loop by selectively querying a house with information deemed essential by our model from the pool set. Subsequently, we deploy a sensor to gather data from this specific house and incorporate the acquired data into our training process. This helps enhance the predictive capabilities of our model by choosing houses with information crucial for better disaggregation.

We employ two active learning settings, one for understanding the upper bounds in the performance (\emph{Query Singly}) and the other, a more practical one (\emph{Query all at once}) active learning respectively.


\subsubsection{Query Singly Active Learning}


In query singly active learning, we individually select a house from the pool set for each appliance. For instance, the house queried for air conditioner could be $h_{5}$ and that for furnace could be $h_{9}$. The selection of a house for an appliance $a_{i}$ is solely based on the uncertainty of that appliance, disregarding other appliances. The acquisition happens according to the equation \ref{query} above.


Uncertainty is assessed using entropy and mutual information to determine the house to be queried. \textbf{However, in practical scenarios, choosing different houses for different appliances is expensive and would require more sensors to be installed. To overcome this limitation, we further investigate and expand upon this approach in the \emph{Query all at once} active learning setting.}



\subsubsection{Query all at once Active Learning}\label{query_at_once}

In "Query All at Once" active learning, we select a single house from the pool set, taking into account all appliances. The selection of the house is based on the weighted uncertainties associated with each appliance. 

\noindent We employ the following query strategies to query a single house across different appliances:\\
\noindent \textbf{1. Uniform Weighted Uncertainty}: The uncertainty is weighted uniformly across all the appliances and the house with the maximum uncertainty is chosen. Thus, the acquisition is modified as: 
\begin{equation}
A\left(\boldsymbol{x}\right)=\sum_{m=1}^{M}{\operatorname{\alpha_m}\left(x \right)* w_m}
\end{equation} \\
\noindent where $M$ is the number of appliances, $a_m(x)$ is the acquisition function and $w_m = \frac{1}{M}$ is the weight associated with the $m^{th}$ appliance.\\
\noindent \textbf{2. \ Rank Sampling}: The uncertainty and thus the acquisition function for each appliance could be on a different scale. To account for the scale variability, we can assign a home a rank as per the acquisition functions for the different appliances and then sum these ranks. Let's consider an example with three pool houses $h_1, h_2, h_3$ and two appliances $a_1, a_2$. Let the corresponding acquisition function values across houses and appliances be $6, 100, 0.03$ and $4, 105, 0.02$. The ranks would be $1,2,1$ and $2,1,2.$ We query $a_1$, based on summation of ranks. 

The house with the highest uncertainty/lower rank considering all appliances is chosen to query for active learning.\\
\noindent \textbf{3. Round Robin Sampling}: In a round robin sampling scheme, we cycle through the acquisition functions for the different appliances and choose this as the main acquisition function. For example, for the first iteration, our acquisition function could be based on air-conditioner, for next iteration it could be furnace, etc. Such a scheme is designed to promote fairness amongst the different appliances. 
\subsection{Pool Temporal Context: Static vs Dynamic}\label{static_vs_dynamic}
Our NILM problem has a special time series flavour to it, making it different from active learning on i.i.d. data setting like images. For example, let us refer back to figure~\ref{problem_statement}. Our initial model is trained on timestamps from $t_1$ (say $1st$ March) through $t_2$ (say $10th$ March). Let us consider a pool home, say $h_6$ and pick a time window, $t_1$ ($1st$  March) to $t_3$ ($15th$ March), for which we predict the uncertainty over the appliance power. This timeframe may have multiple datapoints over which we compute the uncertainty in power drawn over. For example: $1st$ March $8$ AM, $1st$ March $8:30$ AM, $\cdots$ , $15th$ March $11:30$ PM. However, we would need a single metric to capture the uncertainty over the entire time period. We thus make two important considerations: i) the aggregation function; ii) aggregation time window. 

\begin{figure}[h]
\centering
\includegraphics[width=0.9\columnwidth]{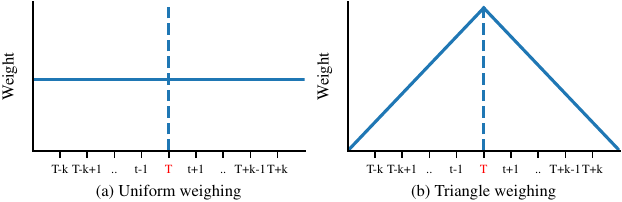}
\caption{Two aggregate function choices: a) uniform, b) triangle}
\label{fig:weighing}
\end{figure}

\begin{figure}[h]
\centering
\includegraphics[width=0.9\columnwidth]{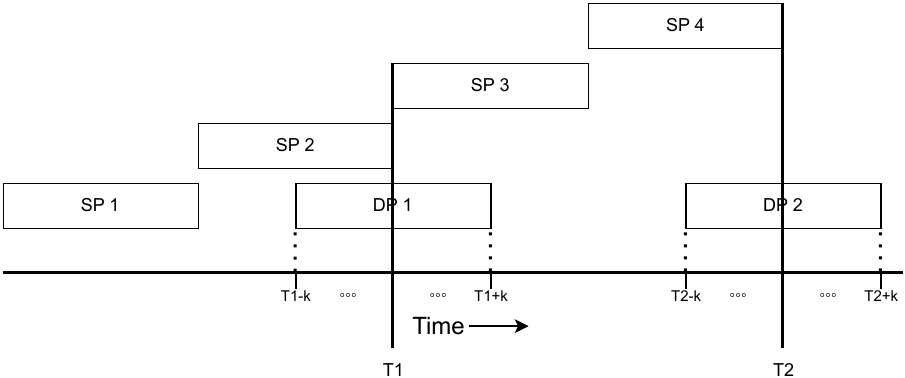}
\caption{Different aggregation time window choices: Dynamic $\mathrm{DP1}$ for $\mathrm{T1}$ and $\mathrm{DP2}$ for $\mathrm{T2}$ or any static window ($\mathrm{SP1, SP2, SP3, SP4}$).}
\label{fig:windows}
\end{figure}

\noindent We now discuss these two considerations in details:

\noindent \textbf{1. \ Aggregation function}: Given a time window $T$ consisting of multiple timestamps, we could aggregate the uncertainty across these timestamps in two ways : i) uniformly by giving equal importance to all the timestamps; or ii) by creating a biased weighing scheme. For a biased weighing approach, we propose a triangle kernel based weighing scheme as shown in figure \ref{fig:weighing}. The weights associated for the uncertainties of the timestamps decrease as we move away from the current timestamp. This helps weigh the current timestamp uncertainty more than that of its neighbouring timestamps and can help choose more ``temporally'' relevant houses.\\
\noindent \textbf{2. \ Aggregation time window}: There can be various choices of time windows over which we can compute the aggregation over the uncertainty. As an example, we could consider a distant past, near future, etc. If we fix the time window over which we compute the aggregation, irrespective of the current timestamp (or iteration of active learning), we call it the static pool. However, we could also consider a dynamic pool window of $K$ days prior to and after the current date to accommodate the temporal information. In figure~\ref{fig:windows}, we show two timestamps ($\mathrm{T1}$) and ($\mathrm{T2}$) where we wish to query pool houses. We could use any of the static pool windows ($\mathrm{SP1, SP2, SP3, SP4}$) or use $\mathrm{DP1}$ dynamic pool window for $\mathrm{T1}$ and $\mathrm{DP2}$ for $\mathrm{T2}$.

\section{Evaluation} \label{eval}
This section discusses the experimental setup used for queries mentioned in the introduction. \textbf{Our work is fully reproducible, and the code is available in our repository.\footnote{\url{https://github.com/anonn23/anon.git}}} 

\subsection{Dataset}

The Pecan Street Dataport database \cite{7418187} is the world's largest residential energy data source. Out of the 725 homes, 75 homes dataset is free for research purposes. Out of these 75 homes, we use the $25$-home dataset from the publicly available dataset of Pecan Street neighborhood of Austin, Texas. The remaining 50 homes have far less data for thorough analysis. 

The dataset includes information on energy consumption, energy generation (from solar panels), appliance-level energy usage, weather data, and demographic information.  We use the one-minute frequency dataset to conduct our analyses and investigations.
Similar to previous studies \cite{bansal2022not, kelly2015neural}, we center our attention on five widely used appliances worldwide: air conditioners, dishwashers, clothes washers, refrigerators, and furnaces. Other appliances have far less data and most families do not have access to it. Additionaly, the dishwasher requires human operation and is occasionally used; on the other hand, the refrigerator operates without any human interaction and might be regarded as a background appliance. Moreover, devices like dishwasher and clotheswasher frequently function in different modes (drying, heating, etc..) and consume varied amounts of power for these different modes.

\subsection{Metrics}
 We use the root mean squared error \cite{10.5555/2900728.2900780} defined as 
\begin{equation}
    RMSE = {\sqrt{\frac{\sum_{i=1}^{n}(y_{i}-\hat{y_{i}})^2}{n}}}
\end{equation}
 
Here, $n$ is the number of samples, $\hat{y_i}$ and $y_i$ is the predicted appliance reading and ground truth reading of an appliance respectively.
Lower RMSE means better performance.

\subsection{Experimental Setup}

The power consumption distribution of certain appliances varies with the seasons. For instance, the air conditioner, predominantly utilized during summer and sparingly during other seasons, exhibits distinct usage patterns based on the seasons. Thus, to make the problem setup realistic, we select our experimental dates from $1^{st}$ March $2018$ to $10^{th}$ May $2018$. During the initial few days, there is zero to minimal air conditioner usage, and the latter days, there is heavy air conditioner usage.

\subsubsection{Disaggregation methods}
We employ sequence-to-point models \cite{zhang2018sequence} and draw inspiration for our hyperparameter selection from earlier studies \cite{batra2014nilmtk}. Due to space constraints, we link the list of hyperparameters here.\footnote{\url{https://github.com/anonn23/anon/blob/main/hyperparameters.md}} 
\begin{enumerate}[leftmargin=*]
    \item Single Output: The input to our S2P model is a $99$-length mains power reading while the output is the mean power prediction of a single appliance and its standard deviation.
    \item Multi Output: The input sequence length is the same $99$-length mains power reading but the output consists of ten nodes; five appliance specific power reading predictions and their respective standard deviations. 
\end{enumerate}

\subsubsection{Train/Pool/Test Split}
We divide the dataset into three parts - five training, five testing and ten pool houses. We select the following house IDs for the train and test split based on prior literature \cite{10.1145/3486611.3486652}. The remaining houses are considered in the pool dataset.
\begin{itemize}
    \item Train indices: $2361, 7719, 9019, 2335, 7951$ 
    \item Test indices:  $4373, 7901, 3456, 3538, 2818$ 
    \item Pool indices: $5746, 8565, 9278, 8156, 8386,\\ 9160, 661, 1642, 7536, 7800$
\end{itemize}

\subsubsection{Base Training}
The active learning process initiates with training the model on five training houses from the set $H_{train}$. We use the aggregate power time series $P_{h}^M(t)$ and the appliance power time series $P_{h_p}^a(t)$ to train the model. The timestamp of training consists of a ten day window, specifically, $1-10$ March $2018$. 

\subsubsection{Active Learning Loop}
\textbf{In each subsequent active learning iteration, we pick the most uncertain pool house and add its aggregate power time series $P_{h}^M(t)$ along with the ground truth data ($P_{h_p}^a(t)$) at an interval of every five days, starting from $11$ March $2018$.}

\subsubsection{Testing Phase}\label{testing_phase}
The model's performance is evaluated on a test set $H_{test}$ ($1-10$ May $2018$), for which the appliance power time series is inaccessible.

\subsubsection{Active learning baselines}
\noindent We propose two active learning baselines namely:\\
\noindent \textbf{Random Baseline}: The random baseline approach involves the selection of houses at random in each iteration. \textbf{We choose random query strategy as the baseline. Majority of the active learning literature uses random sampling as the main baseline for comparison against proposed work. \cite{schohn2000less, settles2009active, cohn1996active, tsymbalov2018dropout}}. Using this acquisition
function is equivalent to choosing a house uniformly at
random from the pool. To mitigate potential biases, we conduct ten repetitions of this experiment, employing different random seeds.\\
\noindent \textbf{Total Baseline}: We train the model on data from train + pool houses from $1$ March - $30$ April $2018$ i.e i.e. we assume that the sensors have been installed in all the pool houses from $1$ March $2018$. The test setup is the same as mentioned in section \ref{testing_phase}.


\subsubsection{Pool Temporal Setup}
To compare between the static and dynamic pool windows discussed in section \ref{stat_vs_dynamic}, we use two static aggregation time window and a dynamic time window, all of fifteen days. The dynamic time window includes seven days in the past, the current day ($T$) (when we evaluate uncertainty over pool set) and seven days in the future. We can thus write the weight from the triangle kernel (defined at current day $T$) for any day $t \in \{T-7, \cdots ,T+7\}$ as:
\begin{equation}
    w_T(t) = 1 - \frac{|T-t|}{8}
\end{equation}
The above formula ensures that our weights vary from $0.125$ at extremes to $1$ at center. In the experiments, the dynamic window starts from $5-20$ March $2018$ and slides forward five days in each subsequent iteration of active learning. The static windows are $16-30$ January $2018$ and $16-30$ April $2018$. We chose the above mentioned static window frames to account for the seasonal change of energy consumption across appliances. We used both the uniform and triangle kernel for the static and dynamic pool windows.

It is important to note that we will never have access to ``future'' data for evaluating pool uncertainty. For the purposes of a practical implementation, one could use the data from the previous year (if available) or introduce a lag in adding pool data (till the end of the pool window). In this paper, we could not leverage data from the previous year due to the dataset limitations. However, we believe such a system to be viable as the mains data is generally easier to collect via the smart metering infrastructure.

\subsubsection{System Settings and Reproducibility}
We use $4$ X NVidia A100 GPUs for training our models, and the code base is written using JAX and Flax libraries. 

\subsection{Results and analysis}
We now present our main results. 
\subsubsection{Query Singly}

\begin{table}[ht]
\centering
\label{sensors_query_singly}
\begin{tabular}{lccccccc}
\toprule
Appliance & \% & MI & Entropy & Random \\
\midrule
\multirow{3}{*}{Air conditioner} & 10\% & 6 & 7 & N.A. \\
& 20\% & 5 & 5 & N.A.\\
& 25\% & 5 & 5 & 10 \\
\midrule
\multirow{3}{*}{Dishwasher} & 10\% & 7 & 8 & N.A.\\
& 20\% & 7 & 7 & N.A.\\
& 25\% & 7 & 7 & N.A.\\
\midrule
\multirow{3}{*}{Clotheswasher} & 10\% & 5 & 8 & N.A.\\
& 20\% & 5 & 8 & N.A.\\
& 25\% & 2 & 5 & N.A.\\

\bottomrule
\end{tabular}
\caption{The table shows the number of sensors required to achieve 10\%, 20\% and 25\% of the total baseline RMSE in different acquisition functions for query singly strategy. `N.A.' refers to not achievable using all the 10 sensors.}
\end{table}

\begin{figure}[h]
\centering
\includegraphics[width=1\columnwidth]{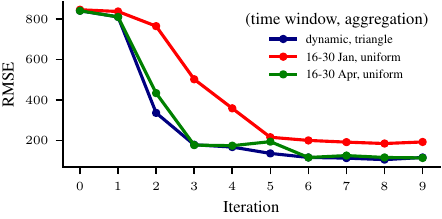}
\caption{Model's performance for air conditioner on two different static windows and a dynamic window with the different aggregation schemes. Dynamic pool window leverages temporal context. As the seasons change and the data becomes available (March-April), performance improves.}
\label{fig:dynamic-air}
\end{figure}

\begin{figure}[h]
\centering
\includegraphics[width=1\columnwidth]{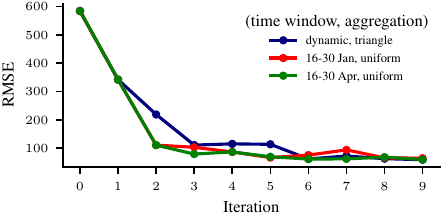}
\caption{Model's performance for dishwasher on two different static windows and a dynamic window with the different aggregation schemes. As the dishwasher is a sparsely used appliance and has no regular patterns, leveraging any temporal information is insignificant.}
\label{fig:dynamic-dishwasher}
\end{figure}

\begin{itemize}[leftmargin=*]
    \item \textbf{AL v/s Random}:
    In figure \ref{single_task_combined}, we see that the active learning approaches using entropy and MI perform better as compared to random approach in almost all the iterations. In addition to that, the errors achieved after the last iteration are comparable with the total baseline error. In other domains \cite{yoo2019learning}, a 3\% improvement over the mean of the random baseline has been considered as state-of-the-art. In contrast, we are well below the sigma interval (68\%) of the random baseline.  
    
    \textbf{Key Analysis}: 
    \begin{itemize}[leftmargin=*]
        \item For air conditioner, we observe that we are able to decrease the errors by about \emph{\textbf{$2x$}} (from \emph{$\sim700$} to \emph{$\sim300$} in the second iteration), just by strategically querying data. 
        \item For air conditioner, we achieve errors closer to the total baseline with six iterations (i.e. with only six new sensors installed in the pool houses as compared to ten new sensors needed to be installed for the total baseline). In case of the dishwasher and clotheswasher, we achieve this performance with just three iterations i.e. three sensors installed on three pool houses using active learning as compared to ten sensors in ten houses in total baseline.
        \item For furnace, we are able to get comparable performance to total baseline in just six iterations of active learning. 
        \item For both clotheswasher and dishwasher, we observe that performance based on MI acquisition function works better than entropy acquisition function, as explained in Section \ref{mc_vs_mi}, as MI considers epistemic uncertainty only.
        \item For refrigerator, all active learning methods exhibit comparable performance. Given its consistent usage throughout the year and nearly uniform energy consumption, the information required for our model remains relatively unaffected by the selection of a house via entropy, MI or the random querying strategy.
        \item 
        Because we use only $\sim65\%$ \footnote{\url{https://github.com/anonn23/anon/blob/main/data.md}}of the data, we may not be able to always match the performance of the total baseline. This is because appliances like refrigerator are continuously used and would require a large amount of data for accurate predictions.
       
    \end{itemize}

    \item \textbf{Entropy v/s MI}\label{mc_vs_mi}:
    As discussed in section ~\ref{mutual information}, differential entropy takes aleatoric (data) and epistemic (model) uncertainty into consideration, MI takes into account only the model uncertainty. \textbf{In sparse appliances like dishwasher and clotheswasher, data uncertainty might be high due to lesser availability of data and inherent stochasticity due to usage.}
    Querying based on differential entropy may result into picking houses that are less informative for our model. Figure \ref{single_task_combined} show that using MI, our models are able to achieve an RMSE of \emph{$100$} with just \emph{$\sim49.3\%$} of training data. Thus, querying using MI is more cost-efficient in sparse appliances.

\begin{figure*}
  \centering
  \includegraphics[scale = 0.6]{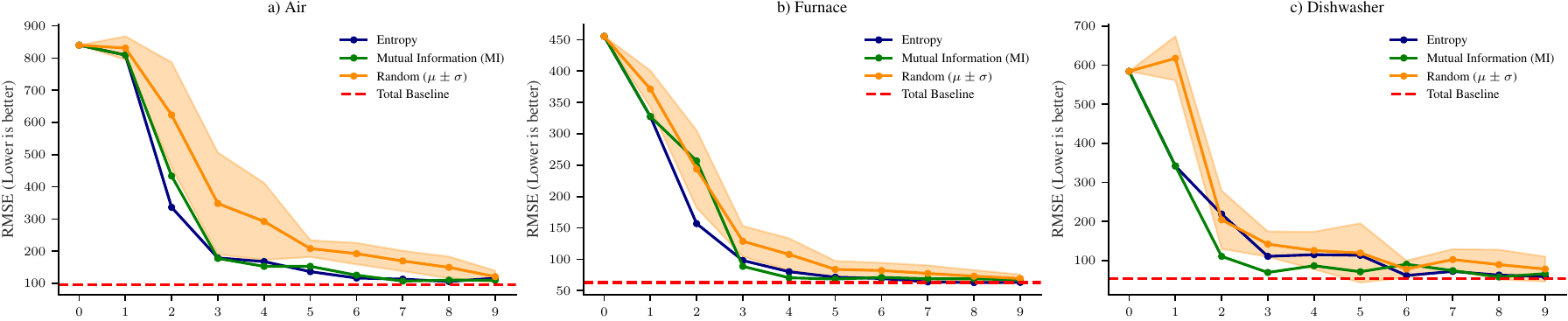}
\end{figure*}
\begin{figure*}
  \centering
  \includegraphics[scale = 0.6]{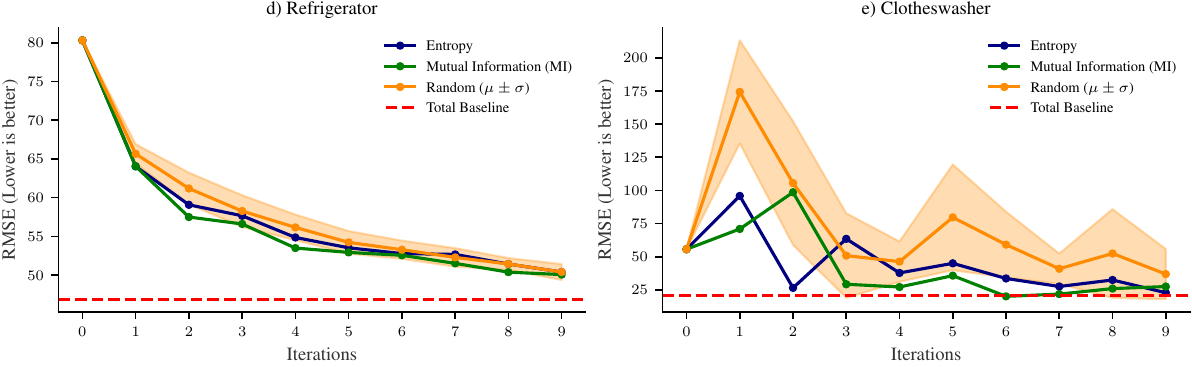}
  \caption{From figure a-e, we observe entropy and MI acquisition functions to perform better than the random baseline for query singly.}
  \label{single_task_combined}
\end{figure*}

    
    \item \textbf{Static V/s Dynamic Pool} \label{stat_vs_dynamic}:
    As stated in section \ref{static_vs_dynamic}, we use static and dynamic pool. 
    In most of the appliances, the dynamic pool achieves errors comparable to the best static error curves possible. For sparsely-used appliances like dishwasher, temporal information is not available for longer periods, due to their irregular usage patterns. Thus, dynamically changing the time periods for pool does not improve performance. 
    However, in frequently-used appliances like air conditioner, temporal information is quite useful as the usage pattern heavily varies according to seasonal change. This allows the model to be more accurate in predicting the power consumption values when the usage pattern changes. We show the results for the two appliances in figures \ref{fig:dynamic-air} and \ref{fig:dynamic-dishwasher} and due to space constraints, have linked the rest results here\footnote{\url{https://github.com/anonn23/anon/blob/main/s_d.md}}. For rest of the experiments, we use dynamic pool with triangle kernel. 
    
    \item \textbf{Analysis}: While query singly active learning gives the best performance, installing sensors at different houses for different appliances is not feasible. It is practical only if the intersection of houses queried across all the appliances is high. Figure \ref{single_task_combined}f shows the number of common houses queried across all the tasks in each iteration. We observe zero common points until the second iteration, and the intersection gradually increases. This shows each task performs independently. We observe that different tasks prioritize different houses in the pool dataset. It is rarely observed that a single house achieves the highest uncertainty across all the tasks. 
    
\end{itemize}

\begin{figure}
  \centering
  \includegraphics{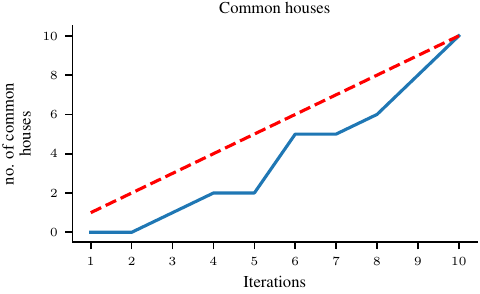}
  \caption{The figure illustrates the number of houses intersected when employing a query singly strategy across all five appliances during active learning iterations. Since the count of shared houses is low, the query singly strategy, which involves querying distinct houses for each appliance, is inefficient.}
  \label{common houses }
\end{figure}

\subsubsection{Query all at once}

\textbf{Due to resource constraints, we have not evaluated the total baseline error for query all at once AL.} 



\begin{figure}[h]
\centering
\includegraphics[width=1\columnwidth]
{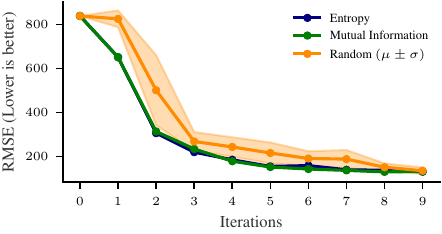}
\caption{Performance analysis of our model on air conditioner in the query all at once active learning setting.}
\label{air_mt}
\end{figure}

\begin{itemize}[leftmargin=*]
    \item \textbf{Active learning v/s Random}

\indent In query all at once active learning, we compute the uncertainties across all appliances during the querying phase, and we query the house as per the different acquisition strategies discussed in section~ \ref{query_at_once}.

We show our results for the uniform weighted uncertainty sampling and attach the  rank and round robin based active learning results in Section \ref{rank} and \ref{round robin}. Figures \ref{air_mt} - \ref{fig:clothes_mt} shows the performance of our query at once active learning.\\ \\
\item \textbf{Analysis}:
\begin{itemize}[leftmargin=*]
    \item MI and entropy perform better than the random acquisition strategy for all appliances across most iterations. We (Entropy, MI) are able to achieve a similar level of performance for air conditioner using only $4$ sensors as compared to the random acquisition function which requires $8$ sensors. Similar inferences can be made for other appliances. 
    \item MI performs distinctively better than entropy for clotheswasher as already observed in query singly settings.  
    \item After the error convergence for query all at once, we observe that the performance of multi-output model is \textbf{slightly worse} than the single output models for query single active learning. For example, after $9$ iterations in air-conditioner the RMSE value for query singly is \emph{$105$} (figure \ref{single_task_combined}a) whereas it is \emph{$136$} (figure \ref{air_mt}) for query all at once. This is due to the query singly strategy optimizing querying for a single appliance (and thus impractical) as opposed to the query all at once strategy optimizing querying for all appliances.
  
    \item The number of iterations required for error-convergence are greater than the query singly active learning for all appliances, for reasons described in the previous point.
\end{itemize}


\end{itemize}

\begin{figure}[h]
\centering
\includegraphics[width=1\columnwidth]
{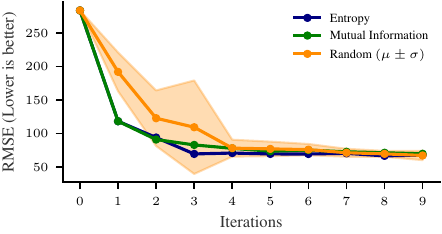}
\caption{Performance of the model on furnace in the query all at once active learning setting.}
\label{fig:example}
\end{figure}



\begin{figure}[h]
\centering
\includegraphics[width=1\columnwidth]
{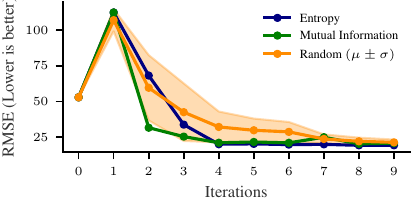}
\caption{Performance of the model on clotheswasher in the query all at once active learning setting.}
\label{fig:clothes_mt}
\end{figure}

\begin{figure}[h]
\centering
\includegraphics[width=1\columnwidth]
{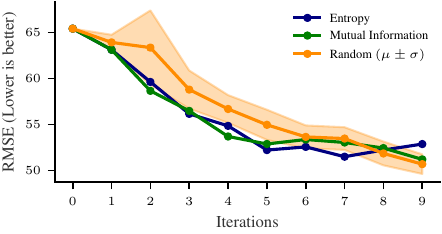}
\caption{Performance of the model on refrigerator in the query all at once active learning setting.}
\label{fig:refrigerator_mt}
\end{figure}

\begin{figure}[h]
\centering
\includegraphics[width=1\columnwidth]
{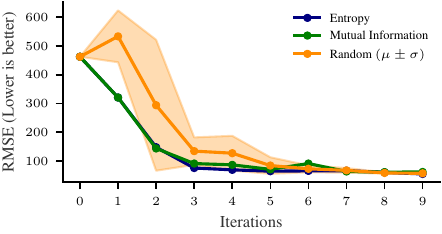}
\caption{Performance of the model on dishwasher in the query all at once active learning setting.}
\label{fig:clothes_mt}
\end{figure}

\section{Rank} \label{rank}

\begin{figure}[H]
\centering
\includegraphics[width=0.9\columnwidth]
{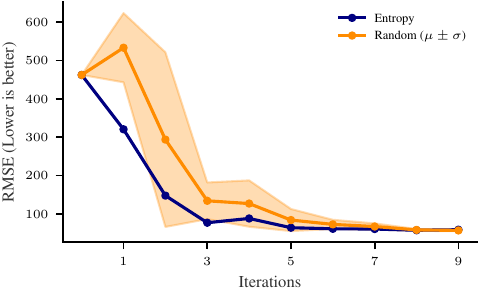}
\caption{Performance of the model on dishwasher in the rank-based active learning setting.}
\end{figure}

\begin{figure}[H]
\centering
\includegraphics[width=1\columnwidth]
{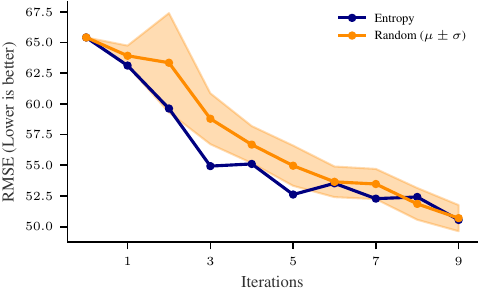}
\caption{Performance of the model on the refrigerator in the rank-based active learning setting. The performance is better than the uniform-weighted active learning}
\end{figure}




\section{Round-Robin} \label{round robin}

\begin{figure}[H]
\centering
\includegraphics[width=1\columnwidth]
{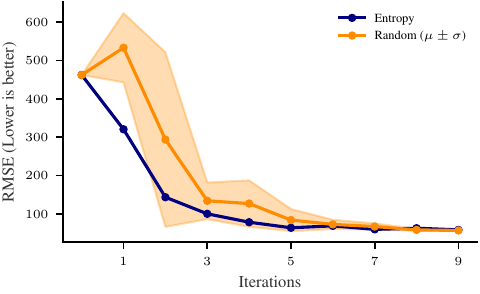}
\caption{Performance of the model on dishwasher in the round-robin active learning setting. Round robin achieves comparable performance to entropy and mutual information.}
\end{figure}

\begin{figure}[h]
\centering
\includegraphics[width=1\columnwidth]
{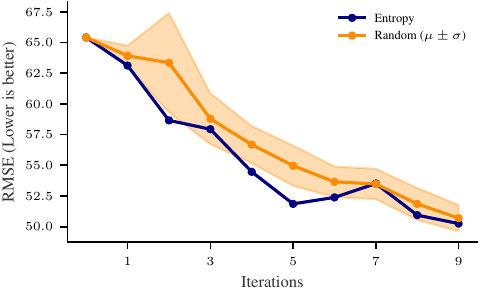}
\caption{Performance of the model on refrigerator in the round-robin active learning setting. Round robin performs better in some iterations as compared to entropy and mutual information.}
\end{figure}




\section{LIMITATIONS, DISCUSSION AND FUTURE
WORK}
\begin{enumerate}[leftmargin=*]
    \item Presently, we do not consider a stopping criteria. In the future, we plan to add heuristic methods to stop when we get diminishing returns. For e.g., we could stop querying once our uncertainty is below a specified threshold. 
    \item In our current approach, we added appliance data from a single home every iteration. However, akin to batch active learning, we could add appliance data from a batch of houses every iteration. Such an approach would require considering the conditional information gain between the different houses chosen in a batch. 
    \item The current approach only looks at data from a single region. In the future, we plan to look into data from multiple regions, where we may need to consider not just the lack of data in a new region, but also the need to fine-tune or transfer the model to a new region.
    
\end{enumerate}




\section{CONCLUSIONS} 
In this paper, we took state-of-the-art NILM models and drastically reduced the requirement for labeled data by strategic querying. Our active learning approach beats the uninformative random sampling strategy. Given the different nature of appliances from a NILM perspective (regularly used like fridge, or sparsely used like dishwasher), we found that acquisitions that account for model uncertainty alone are likely to do better. Overall, we believe that given the non-trivial challenges in acquiring ground truth data in many Buildsys applications, active learning approaches could reduce the labeling and sensor maintenance cost. This work contributes to the benchmarking track by demonstrating the potential of active learning in optimizing NILM systems.  

\newpage
\bibliographystyle{ACM-Reference-Format}
\bibliography{main}

\end{document}